\documentclass[journal]{IEEEtran}
\usepackage{algorithmic}
\usepackage{algorithm}
\usepackage[utf8]{inputenc}
\usepackage[T1]{fontenc}
\usepackage{siunitx}
\usepackage[colorlinks=true, allcolors=blue]{hyperref}
\usepackage{amsmath}
\usepackage{makecell}
\usepackage{amsfonts}
\usepackage{soul} 
\usepackage{subcaption}
\usepackage{stfloats}
\usepackage{booktabs}
\usepackage{enumitem}

\usepackage{tikz,xcolor}
\definecolor{lime}{HTML}{A6CE39}
\DeclareRobustCommand{\orcidicon}
{
    \begin{tikzpicture}
    \draw[lime, fill=lime] (0,0) circle [radius=0.16] 
    node[white] {{\fontfamily{qag}\selectfont \tiny ID}};    \draw[white, fill=white] (-0.0625,0.095) circle [radius=0.007];    
    \end{tikzpicture}
    \hspace{0mm}}
\foreach \x in {A, ..., Z}{%
    \expandafter\xdef\csname orcid\x\endcsname{\noexpand\href{https://orcid.org/\csname orcidauthor\x\endcsname}{\noexpand\orcidicon}}
}

\setlist[enumerate]{itemsep = 0pt, parsep = 0pt, topsep = 0pt} 
\setlist[itemize]{itemsep = 0pt, parsep = 0pt, topsep = 0pt} 
\begin{document}
\title{Exploration of the Assessment for AVP Algorithm Training in Underground Parking Garages Simulation Scenario}

\author{Wenjin Li
}

\markboth{Journal of \LaTeX\ Class Files,~Vol.
}%
{Shell \MakeLowercase{\textit{et al.}}: Bare Demo of IEEEtran.cls for IEEE Journals}

\maketitle

\begin{abstract}
The autonomous valet parking (AVP) functionality in self-driving vehicles is currently capable of handling most simple parking tasks. 
However, further training is necessary to enable the AVP algorithm to adapt to complex scenarios and complete parking tasks in any given situation. 
Training algorithms with real-world data is time-consuming and labour-intensive, and the current state of constructing simulation environments is predominantly manual. This paper introduces an approach to automatically generate 3D underground garage simulation scenarios of varying difficulty levels based on pre-input 2D underground parking structure plans.
\end{abstract}


%
\IEEEpeerreviewmaketitle

\section{Introduction}




Simulation test scenarios are an important part of helping autonomous driving algorithms improve, but current simulation scenarios are still limited to manual approaches. 
The ultimate goal of this project is to generate Autonomous Valet Parking (AVP) simulation test scenarios in underground garages with differentiated difficulty levels through reinforcement learning, which will challenge the vehicle-side AVP algorithms and ultimately improve the algorithmic test metrics. The main work at this stage is to generate a complete 3D underground garage scene with difficult description and differentiation based on a given 2D underground garage plan structure (including only the road and the relative position description of parking spaces).
\par
For the 3D scene modeling, we will use a game engine with a complete weather and lighting system for automated modeling. 
To describe the difficulty of the static simulation scene, we need to define a complete set of evaluation systems associated with it. Given that current parking algorithms are capable of handling parking tasks in simple scenarios, our specific task at this stage is to generate challenging (difficult) underground garage scenarios.

\par
For an AVP vehicle-side system, we want it to be able to accomplish the parking task safely. First of all, we do not want to have accidents such as vehicle collisions during the operation of the system, which is often caused by the perception algorithms in the system failing to recognize the target vehicles or pedestrians correctly at a safe distance.
To avoid such incidents, the vehicle-side system needs to ensure that the perception algorithms can detect other traffic participants around it in a timely and accurate manner as much as possible, in addition to that, we need to make sure that they can accurately recognize lane markings to accomplish the parking task.

\par
The challenge, in reality, is that the presence of objective factors such as building structure occlusion, vehicle occlusion, and lighting arrangement in the environment of an underground garage makes the perception algorithms unable to accurately identify traffic participants who are at risk of crossing around their vehicles in some scenarios. Based on the above, we prioritize at this stage the challenges arising from the objective influences in the underground garage structure on the perception algorithms of the vehicle-side system.





\section{Related Work}
\label{sec:related_work}
Abdullah's study, as described in \cite{Abdullah}, compared the space utilization efficiency of diagonal, parallel, and perpendicular parking methods. The research findings concluded that perpendicular parking methods yield the highest number of parking spaces. This conclusion was drawn using a university as a specific example.
The study summarized in \cite{Lin2017smartparking} focuses on smart parking solutions, emphasizing their significance in the context of urban growth and traffic congestion. 
This research paper meticulously reviews the existing literature, categorizes prior research efforts, elucidates the methodologies employed, and provides practical insights \cite{lan2016development}.
As the number of vehicles continues to increase, the search for available parking spaces has become a growing challenge. The research presented in \cite{KHALID2021survey} underscores the importance of efficient parking solutions, ranging from Smart Parking (SP) to emerging technologies like Autonomous Valet Parking (AVP). 
These solutions encompass digitally enhanced parking, smart routing, and AVP for both short and long distances. 
The survey also highlights potential future research directions and unresolved issues in this field.
Furthermore, the research article by \cite{Holger2017survey} offers an in-depth review of the Autonomous Valet Parking (AVP) field. 
It introduces the core components of AVP, encompassing platforms, sensors, maps, localization, perception, environment modeling, and motion planning. 
Additionally, the article explores the possibilities of High-Density Parking (HDP), which can either reduce the space requirements for parking or increase the capacity of parking facilities. 
The article concludes by emphasizing the remaining challenges and technological prerequisites in both the AVP and HDP domains.
\cite{jimenez2016advanced} presents the ADAS (Advanced Driver Assistance System) that combines artificial vision \cite{lan2022vision,lan2018real}, 3D-laser scanning, and wireless communication.
This system significantly enhances road safety, anticipates potential hazards, and provides efficient driving guidance while issuing warnings and taking control when necessary. 
Extensive testing in controlled environments has yielded positive results.
In the realm of enhancing parking systems, the study detailed in \cite{lim2019radar} introduces a network that combines radar and camera data to improve parking outcomes, even when using auto-annotated data. 
The authors acknowledge the potential for further improvements with better annotation and the exploration of additional scenarios, such as urban driving and adverse weather conditions.

\section{Methodology}
\label{sec:methodology}
\begin{table}[!ht]
\begin{center}
\caption{The symbol table.}
\begin{tabular}{c|c}
\textbf{symbol} & \textbf{definition}\\ \hline
\(\mathcal{S} \) & \makecell{Structure matrix, each element of which represents the\\ type of objects placed in the corresponding area} \\
\(\mathcal{S}(i,j) \) & an element of $\mathcal{S}$ whose coordinates are (i, j) \\
\(\mathcal{R} \) & \makecell{A matrix of row values, each element of which represents\\ the horizontal width of the corresponding area} \\
\(\mathcal{R}(i,j) \) & an element of $\mathcal{R}$ whose coordinates are (i, j) \\
\(\mathcal{C} \)&  \makecell{A matrix of column values, each element of which represents\\ the vertical width of the corresponding area} \\
\(\mathcal{C}(i,j) \) & an element of $\mathcal{C}$ whose coordinates are (i, j) \\
\(m_\delta \) & the number of rows in matrix $\delta$ \\
\(n_\delta \) & the number of columns in matrix $\delta$ \\
\(\eta \) & a square of the garage whose structure matrix is $\mathcal{S}$\\
\(\mathcal{S}(\eta) \) & the element corresponding to $\eta$ \\
\(\theta_{i} \) & Four adjacent squares around square i \\
\end{tabular}
\end{center}
\end{table}
\subsection{2D underground garage structure}
Our automated 3D underground garage scene will use a given 2D underground garage structure as input. We start by defining this underground garage planar structure.
After examining the structure of the underground garage, we find that the layout of the column network of the underground garage is regular, so we can divide the planar structure of the underground garage into regular blocks according to the layout of the column network. Here we use rasterization to divide the underground garage plane and use three different matrices (structural matrix, row matrix, column matrix) to represent the specific information of this plane, where each element of the structural matrix $\mathcal{S}$ represents its corresponding raster plot category, and each element of the row matrix $\mathcal{R}$ and column matrix $\mathcal{C}$ represents the length and width of its corresponding raster plot. With these three matrices, we can store the main structural information of the underground garage to facilitate its 3D modeling in the simulator.
\begin{equation}
\label{deqn_ex1}
\mathcal{S}(i,j)=\left\{
\begin{aligned}
-1&, & & \text{obstacle} \\
0&, & & \text{parking space or free space} \\
1&, & & \text{lane} \\
2&, & & \text{entrance} \\
3&, & & \text{exit} \\
\end{aligned}
\right.
\end{equation}
\subsection{3D simulated underground garage}
We will complete the 3D modeling of the underground garage scene in the following steps:
\subsubsection{Check the reasonableness of the input data}
for the three input matrices, if the size of the structure matrix is $m \times n$ then we need to ensure that the following relationship exists:
\begin{equation}
\label{deqn_ex1}
m_R = m ,\quad n_R = 1
\end{equation}
\begin{equation}
\label{deqn_ex1}
m_C = 1 ,\quad n_C = n
\end{equation}
For the elements in the structure matrix, we must ensure that: 
\begin{equation}
\label{deqn_ex1}
\mathcal{S}(i,j) \in \mathbb{N} ,\quad -1 \leq \mathcal{S}(i,j) \leq 3
\end{equation}
\subsubsection{Completing the structural information of each square based on the raw data}
\begin{figure*}   \centering
    \includegraphics[width=0.95\textwidth]{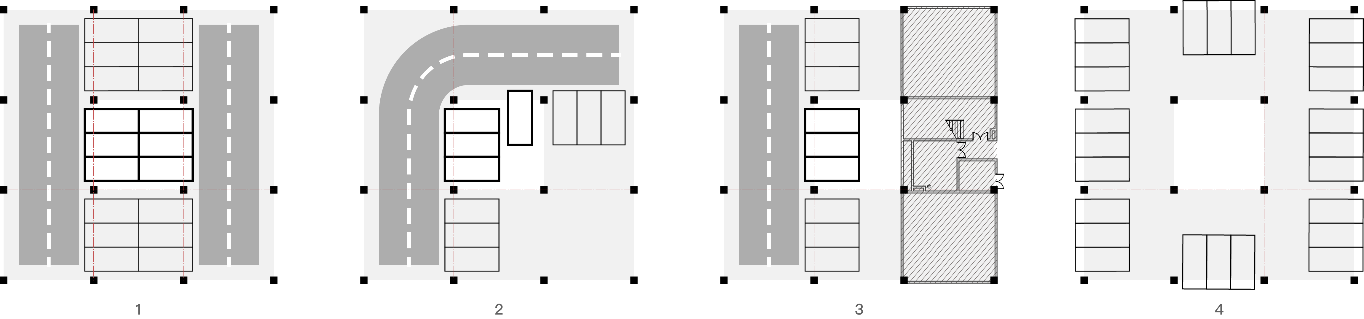}
    \caption{Parking Space Type}
    \label{fig:PST}
\end{figure*}

after ensuring that the data is sound, we need to analyze the raw data even further by storing the information about the analyzed square using a data structure that can contain more information, which includes the spatial location of the square (row and column coordinates) as well as the specific class of the square \cite{lan2022class,gao2021neat}. For example, for a square labelled as a lane, we would further record which of the following four types of roads it belongs to a straight road, a corner road, a T-junction, or a crossroads, and our classification method would satisfy the following equation. 
We first count the number of lane squares which are adjacent to square $\eta$
\begin{algorithm} 
	\caption{count the number of lane squares} 
	\label{alg3} 
	\begin{algorithmic}
    \STATE $cnt = 0$
	\STATE $\theta_\eta \gets GetAdjacentSquares(\eta)$ 
	\FOR{$i$ in $\theta_\eta$}
		\IF{$i$ is lane square}
		    \STATE $cnt$++
		\ENDIF
	\ENDFOR
	\RETURN  $cnt$
	\end{algorithmic} 
\end{algorithm}
and for any square $\eta$, if:
\begin{equation}
\label{deqn_ex1}
\mathcal{S}(\eta) = 1
\end{equation}
we define $\eta$ as:
\begin{equation}
\label{deqn_ex1}
\eta=\left\{
\begin{aligned}
Crossroads&, & & \text{cnt = 4} \\
T-junction&, & & \text{cnt = 3} \\
Straight road&, & & \text{cnt $\leq$ 2} \\
\end{aligned}
\right.
\end{equation}
and if:
\begin{equation}
\label{deqn_ex1}
\mathcal{S}(\eta) = 0
\end{equation}
we define $\eta$ as different types in \autoref{fig:PST}:
\begin{equation}
\label{deqn_ex1}
\eta=\left\{
\begin{aligned}
Type1&, & & \text{cnt $\geq$ 3 or across cnt = 2} \\
Type2&, & & \text{adjacent cnt = 2,} \\
Type3&, & & \text{cnt = 1} \\
Type4&, & & \text{cnt = 0} \\
\end{aligned}
\right.
\end{equation}
\par
In addition, since our models are all loaded in the same orientation in the simulator, we also need to determine the angle of rotation of each square to its adjacent squares and store the angle of rotation in the above square data structure.
\subsubsection{Automate the modeling based on the obtained data}
after adding the specific data of the underground garage structure, we will bind the corresponding models for different squares according to their structural information, and model them in the order of square to column network to ceiling.
\subsubsection{Underground garage details added or modified as required}
To simulate the underground garage in different states, we can add or modify some information to the structure generated above, such as: parked or moving vehicles, light intensity, location of the column network, type and relative location of parking spaces, and type of lanes and their widths. All of the above-listed factors have a direct or indirect effect on the perception of the self-driving vehicle in the underground garage. Because of this, these factors may also serve as criteria for us to evaluate the difficulty of an underground garage.

\subsection{Impact of different factors on perception of autonomous vehicles}
Autonomous vehicles are equipped with a variety of different sensors, such as cameras, LiDAR \cite{xu2019online}. Here we mainly consider the effect of environmental factors on the camera. Firstly, we shall proceed with the categorization of a part of possible influencing factors.
\begin{itemize}
    \item Factors that have a direct effect on perception: light intensity, blurring of lane lines.
    \item Indirectly affecting perception: type and position of column network, colour and size of the vehicle, type and width of the lane.
\end{itemize}
\par
Within this research study, our primary focus centres on the examination of two key variables, namely occlusion and light intensity. We will employ the YOLO (You Only Look Once) algorithm to perform vehicle recognition across various simulated scenarios.
Since all our experiments are conducted in simulation scenarios, to make sure that the model has a high degree of recognition of real vehicles, so that its recognition in simulation scenarios will be meaningful afterwards, we will first test the model with the vehicle dataset. 
If the YOLO pre-trained model is not able to have good recognition, then we will use this dataset to re-train a model for recognition detection in the simulation scenario.

\par
Through multiple iterations utilizing diverse pre-trained YOLO models, a composite confidence level shall be derived as the outcome of the recognition process. Subsequently, we intend to assess the impact of occlusion and light intensity on vehicle recognition across a spectrum of distinct scenarios, outlined as follows:
\subsubsection{Individual testing of the effect of light intensity on vehicle recognition}
four different intensities of light were prepared for the scene in this experiment:
\begin{itemize}
    \item Bright: fully covered and bright light system
    \item Clear: fully covered light system with moderate light intensity
    \item Moderate: mostly covered light system with moderate light intensity
    \item Dim: necessarily covered light system with moderate light intensity
\end{itemize}
We will test the model's ability to recognize different unoccluded vehicles in the same scene under the above four different light intensities.
\subsubsection{Testing the effect of different occlusion scenarios on recognition}
this experiment is to test the effect of occlusion on vehicle recognition alone, while for the occlusion problem in the underground garage scenario, we mainly consider the following cases:
\paragraph{Self-vehicle is travelling in a lane, recognizing a vehicle in the same lane at a corner}
in this case, we will test the effect of a column on the corner of a road on the self-vehicle's ability to recognize the vehicle in front of it that is occluded by the column. The reason for selecting this case is to simulate a rear-end collision due to the failure of the self-vehicle in the lane to recognize the vehicle in front of it.
\paragraph{Self-vehicle parks in a space, recognizing vehicle about to pass in the lane ahead}
in this case, we will test the effect of a column on the side of a parking space on the identification of a vehicle in the lane that is occluded by it. The reason for selecting this case is to simulate an accident in which a self-vehicle in a parking space fails to recognize a vehicle on the road in front of it, resulting in a collision with a vehicle on the road in front of it as it leaves the parking space.
\paragraph{Self-vehicle is driving in a lane, recognizing vehicles in the surrounding parking spaces}
in this case, we will test the effect of a column on the side of a parking space on the identification of vehicles in the parking space that are occluded by it. The reason for selecting this scenario is to simulate an accident in which a self-vehicle in a lane fails to recognize a parked vehicle around the road in front of it, resulting in a collision with a vehicle that is leaving a parking space while it is travelling.

\par
In all the aforementioned scenarios, the timely and accurate recognition of objects with potential cross-path risk by the host vehicle would effectively mitigate the occurrence of accidents, demonstrating the adaptability of the current algorithm to challenging environmental conditions \cite{lan2019evolving}. 
However, should recognition difficulties arise due to the structure's characteristics of that underground garage, leading to accidents, we can subsequently refine the algorithm based on the existing parking structure to enhance its ability to adapt to various challenging environments, ultimately resulting in algorithm improvement.







\section{Experiments}
\label{sec:challenges}
\subsection{Model verification}
\begin{figure*} [!ht]  \centering
    \begin{minipage}{0.30\linewidth}       \centering
      \includegraphics[width=\linewidth]{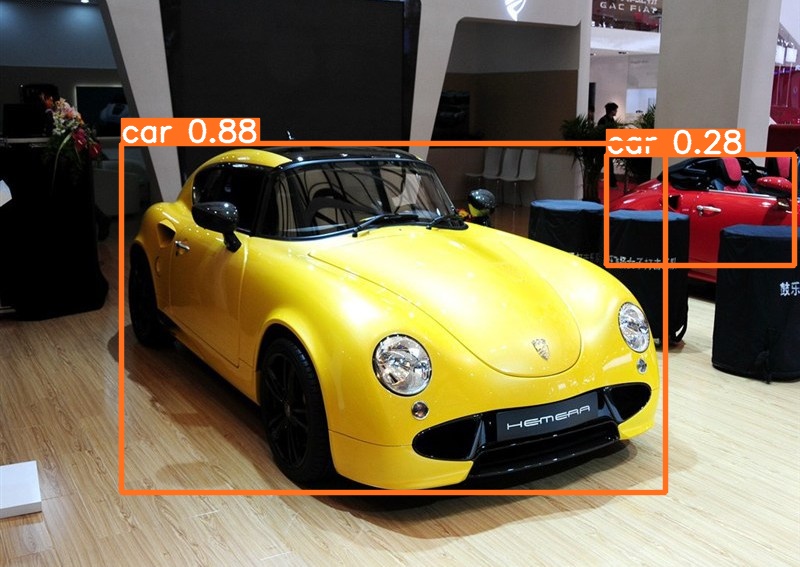}
      \caption{Real Car1}
      \label{fig:realcar_1}
    \end{minipage}
    \begin{minipage}{0.30\linewidth}       \centering
      \includegraphics[width=\linewidth]{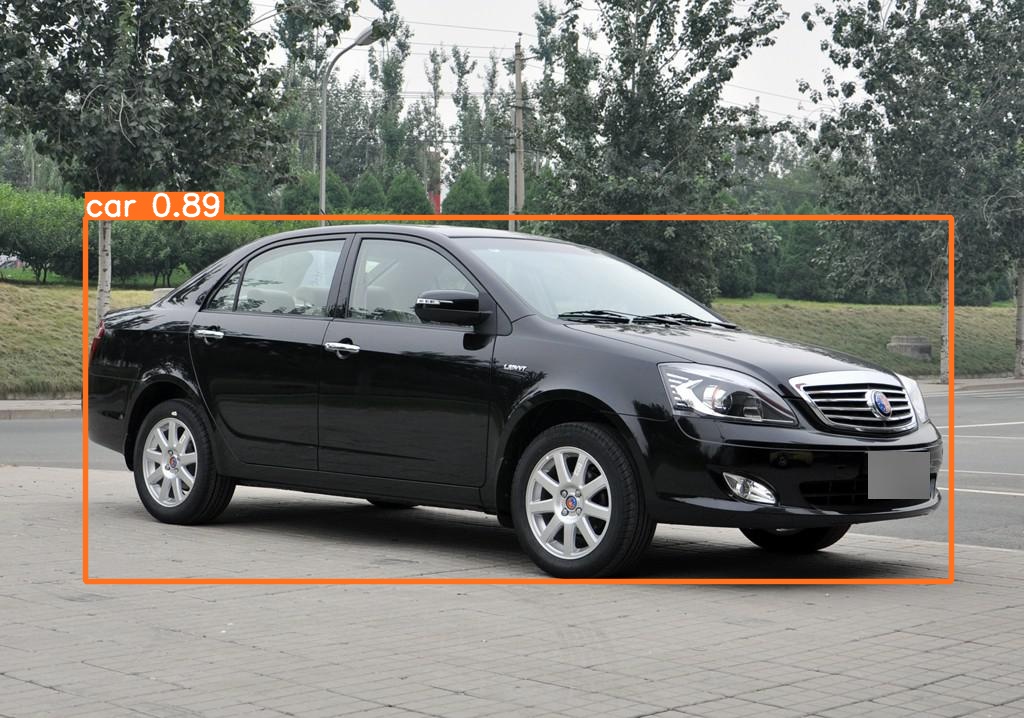}
      \caption{Real Car2}
      \label{fig:realcar_2}
    \end{minipage}
    \begin{minipage}{0.30\linewidth}       \centering
      \includegraphics[width=\linewidth]{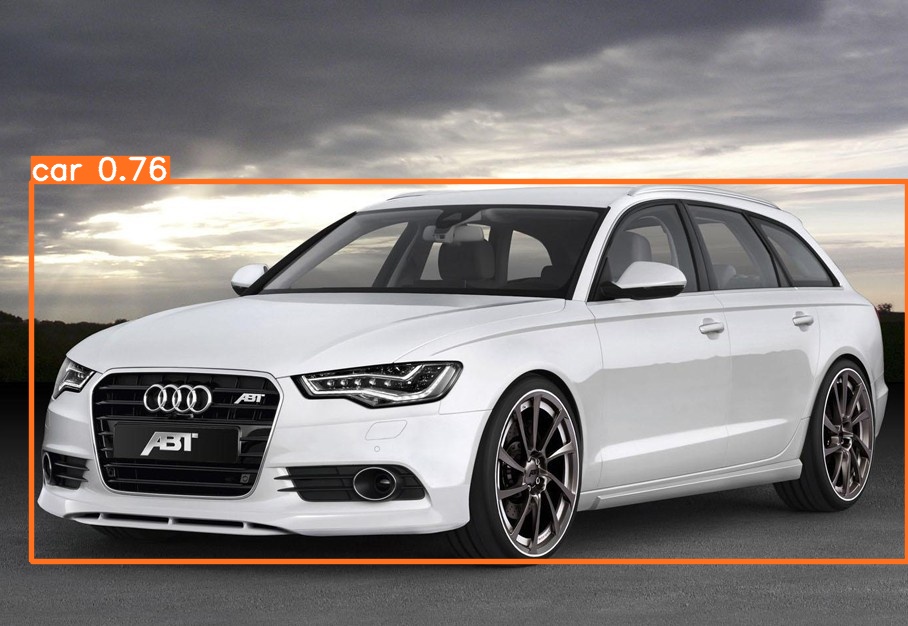}
      \caption{Real Car3}
      \label{fig:realcar_3}
    \end{minipage}
    \hspace{0.05\linewidth}
\end{figure*}

We selected the CompCars dataset\cite{CompCars} for validating the YOLO model, and the partial results (\autoref{fig:realcar_1}, \autoref{fig:realcar_2}, \autoref{fig:realcar_3}) of the validation are as follows. It can be observed that the YOLO model we chose exhibits a high level of recognition confidence when applied to real-world vehicles.
\subsection{Setup}
In the following experiments, we employed the Unity3D engine as our simulator. Regarding image acquisition, we only considered the perspective of a front-facing camera positioned at the top of the vehicle with a field of view (FOV) of 60 degrees.
\subsection{Experiment of light intensity}
\begin{figure*} [!ht]  \centering
    \begin{minipage}{0.47\linewidth}       \centering
      \includegraphics[width=\linewidth]{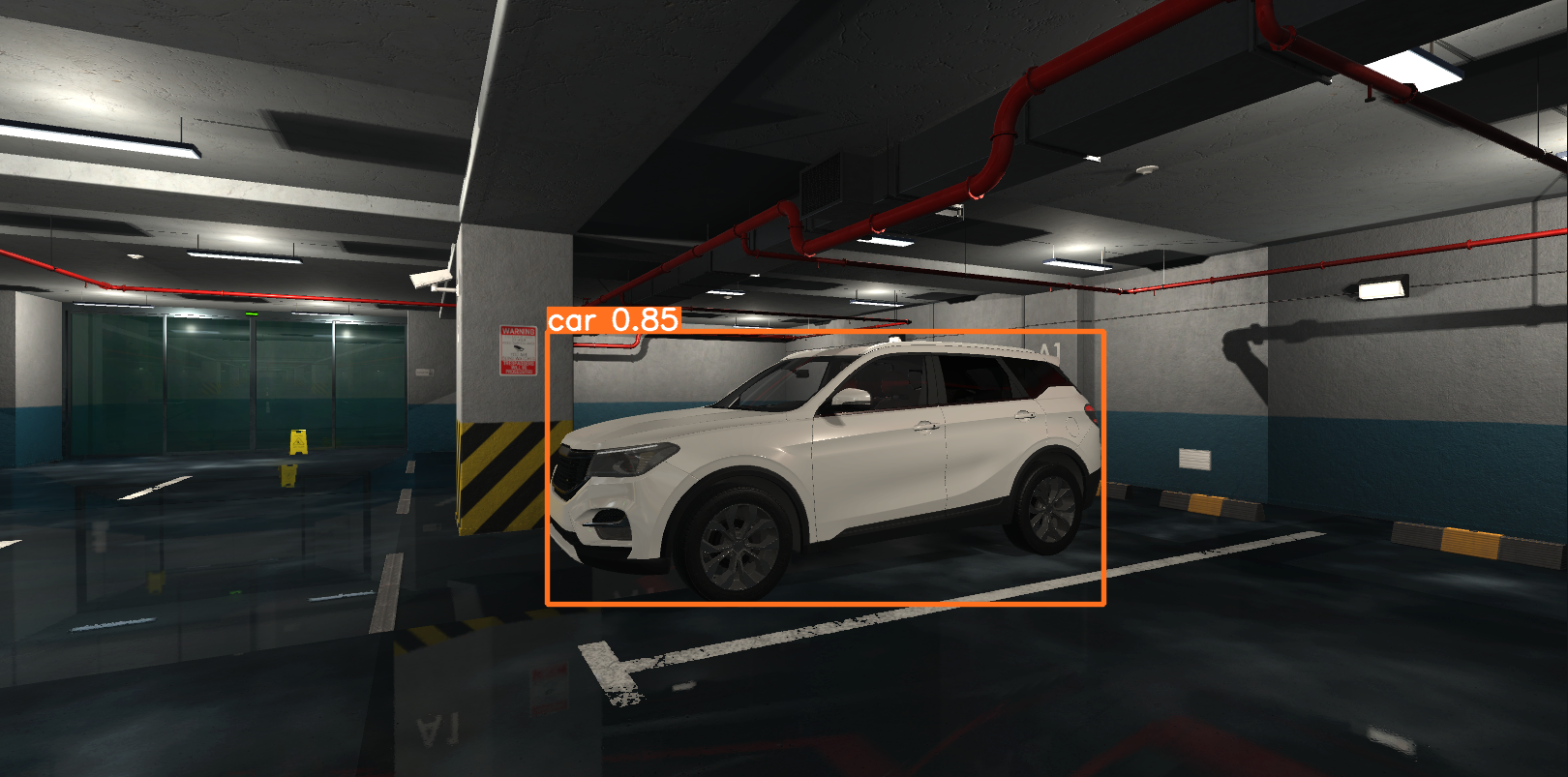}
      \caption{White car in Bright light}
      \label{fig:lit_exp_bright}
    \end{minipage}
    \begin{minipage}{0.47\linewidth}       \centering
      \includegraphics[width=\linewidth]{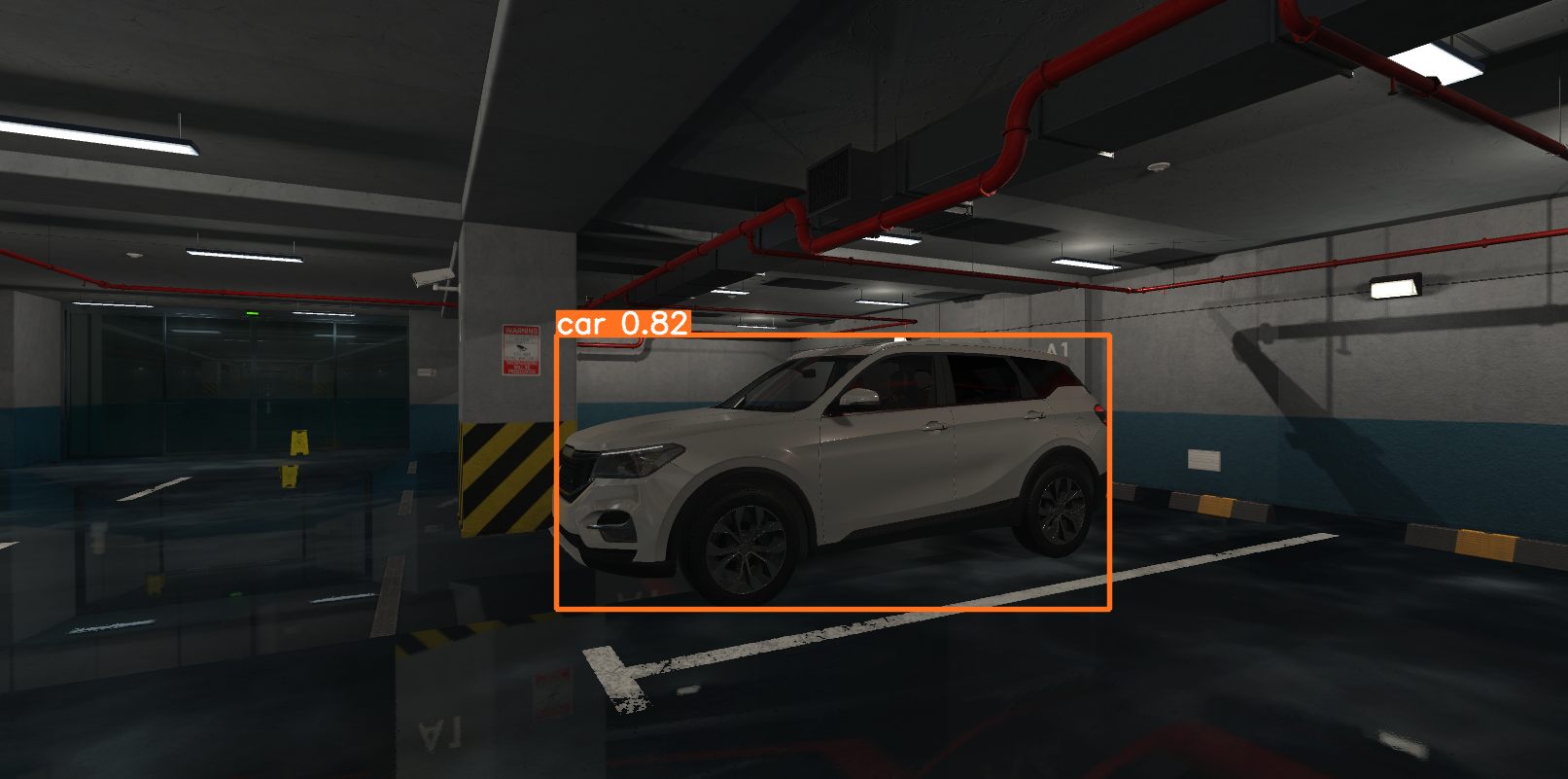}
      \caption{White car in Clear light}
      \label{fig:lit_exp_clear}
    \end{minipage}
    \begin{minipage}{0.47\linewidth}       \centering
      \includegraphics[width=\linewidth]{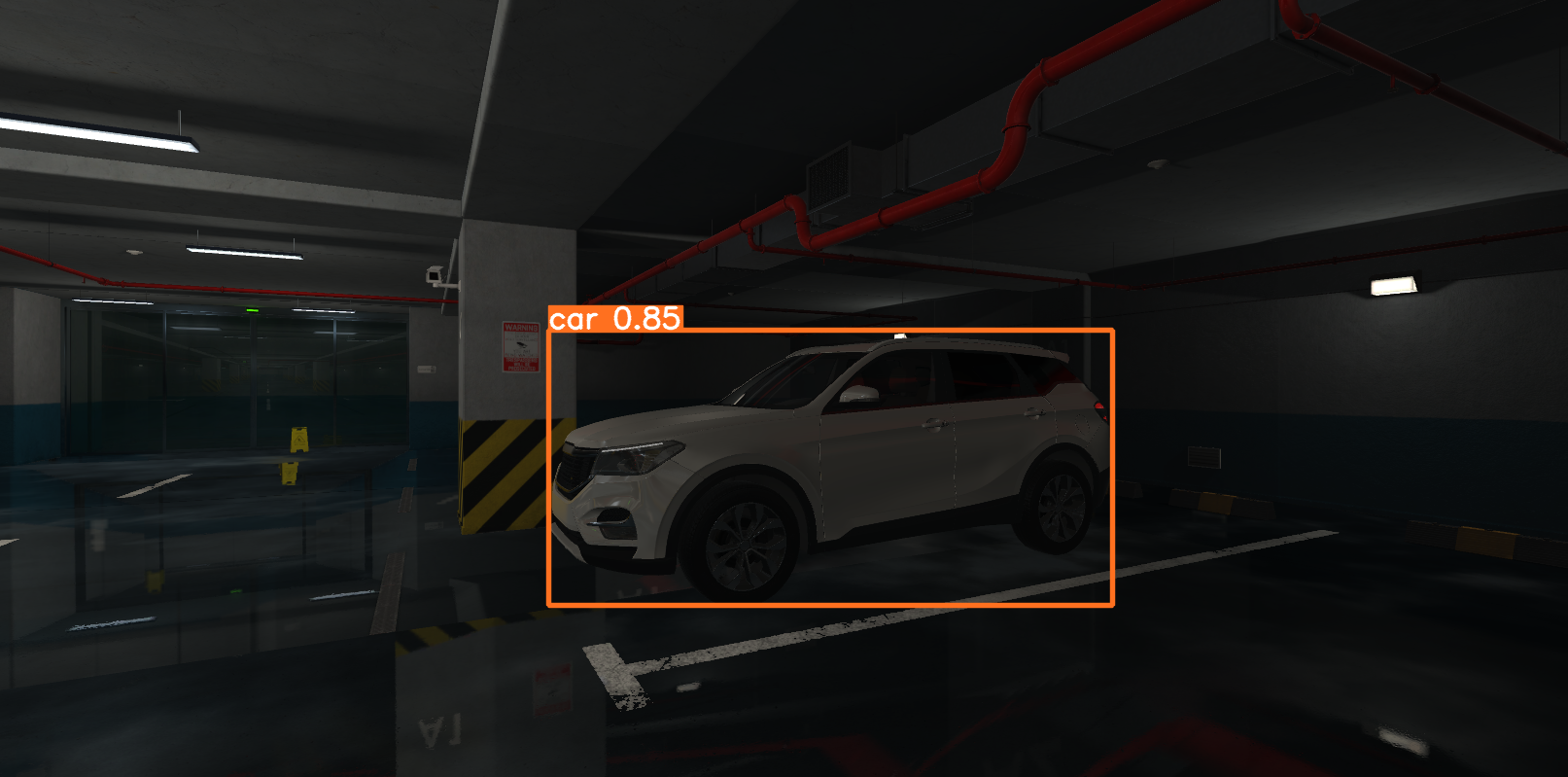}
      \caption{White car in Moderate light}
      \label{fig:lit_exp_mod}
    \end{minipage}
    \begin{minipage}{0.47\linewidth}       \centering
      \includegraphics[width=\linewidth]{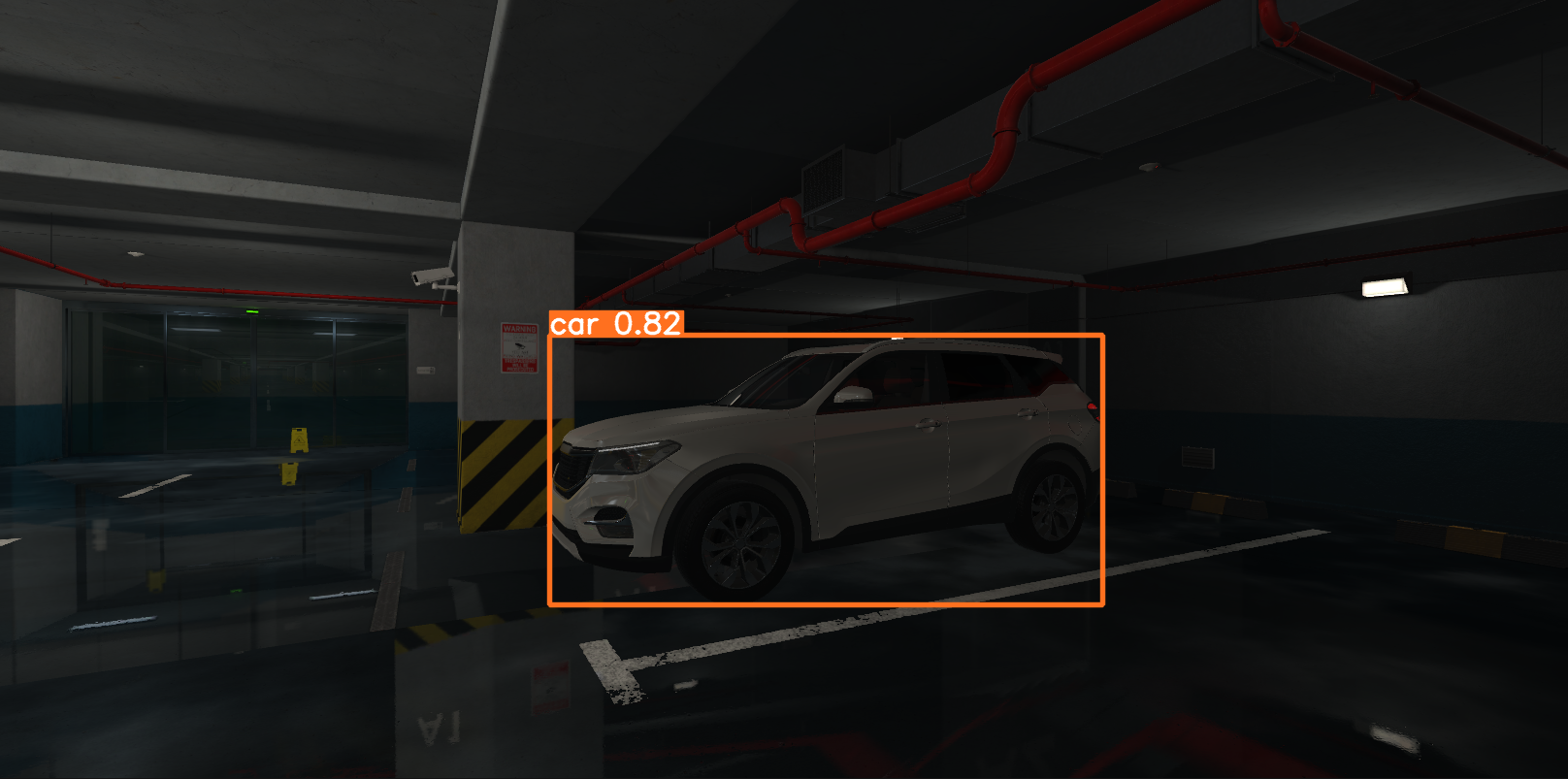}
      \caption{White car in Dim light}
      \label{fig:lit_exp_dim}
    \end{minipage}
    \hspace{0.05\linewidth}
\end{figure*}

We conducted tests on the recognition confidence of vehicles of two different colours (black and white) under varying light intensity conditions within the same scene (\autoref{fig:lit_exp_bright} to \autoref{fig:lit_exp_dim}). In order to eliminate the influence of background on vehicle recognition, we also conducted a control experiment under the condition of background removal. The results are shown in \autoref{tab:light_exp} and \autoref{tab:light_blank}.
\begin{table}
    \centering
    \begin{tabular}{c|cccc} \toprule
         & Bright & Clear & Moderate & Dim\\ \midrule
        Black car confidence level & 0.87 & 0.85 & 0.87 & 0.88\\
        White car confidence level & 0.85 & 0.82 & 0.85 & 0.82\\ \bottomrule
    \end{tabular}
    \caption{Experimental Group Data(Only light)}
    \label{tab:light_exp}
\end{table}

\begin{table}
    \centering
    \begin{tabular}{c|cccc} \toprule
         & Bright & Clear & Moderate & Dim\\ \midrule
        Black car confidence level & 0.90 & 0.91 & 0.88 & 0.88\\
        White car confidence level & 0.90 & 0.91 & 0.90 & 0.90\\ \bottomrule
    \end{tabular}
    \caption{Blank control Group Data(Only light)}
    \label{tab:light_blank}
\end{table}

\subsection{Experiment of occlusion in different situations}
\subsubsection{case one}
\begin{figure*} [!ht]  \centering
    \begin{minipage}{0.47\linewidth}       \centering
      \includegraphics[width=\linewidth]{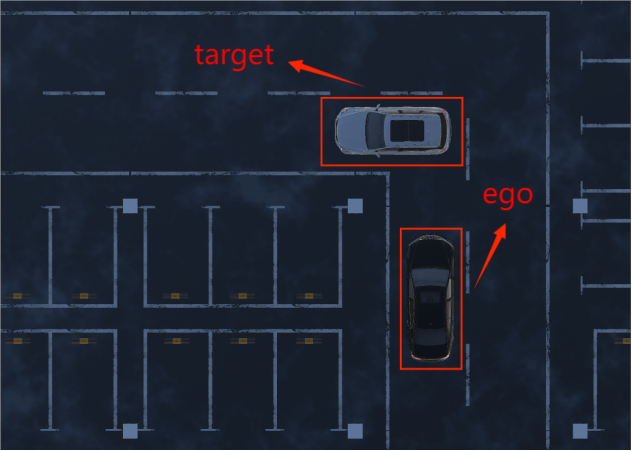}
      \caption{Case1 Start State}
      \label{fig:occ_1_st}
    \end{minipage}
    \begin{minipage}{0.47\linewidth}       \centering
      \includegraphics[width=\linewidth]{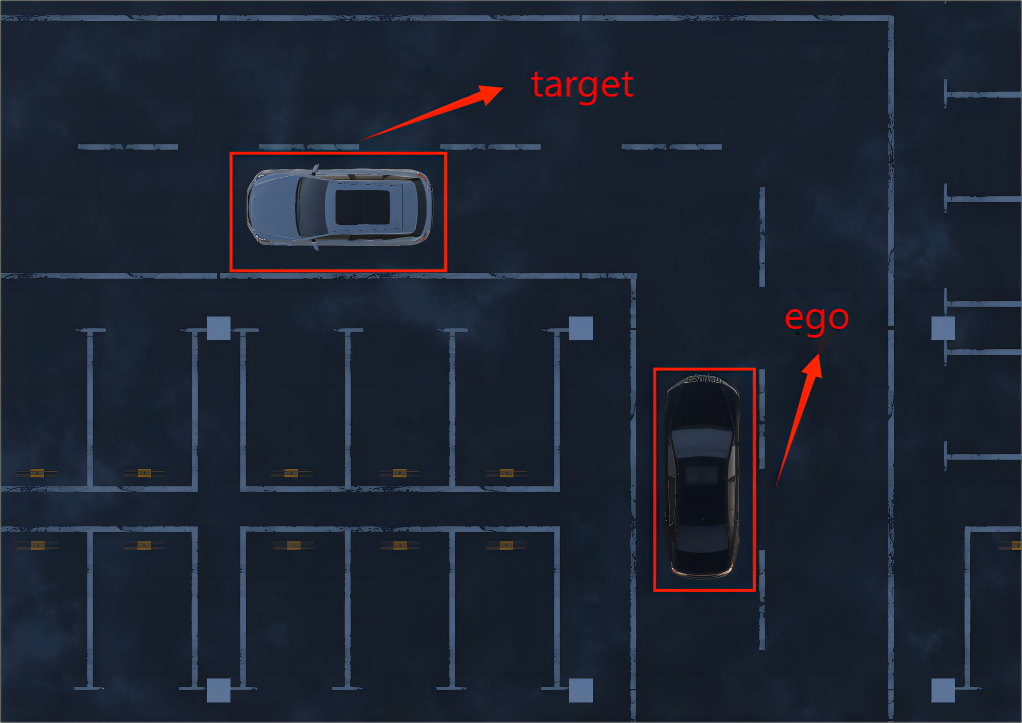}
      \caption{Case1 End State}
      \label{fig:occ_1_ed}
    \end{minipage}
    \hspace{0.05\linewidth}
\end{figure*}

\begin{figure*}   \centering
    \includegraphics[width=0.9\textwidth]{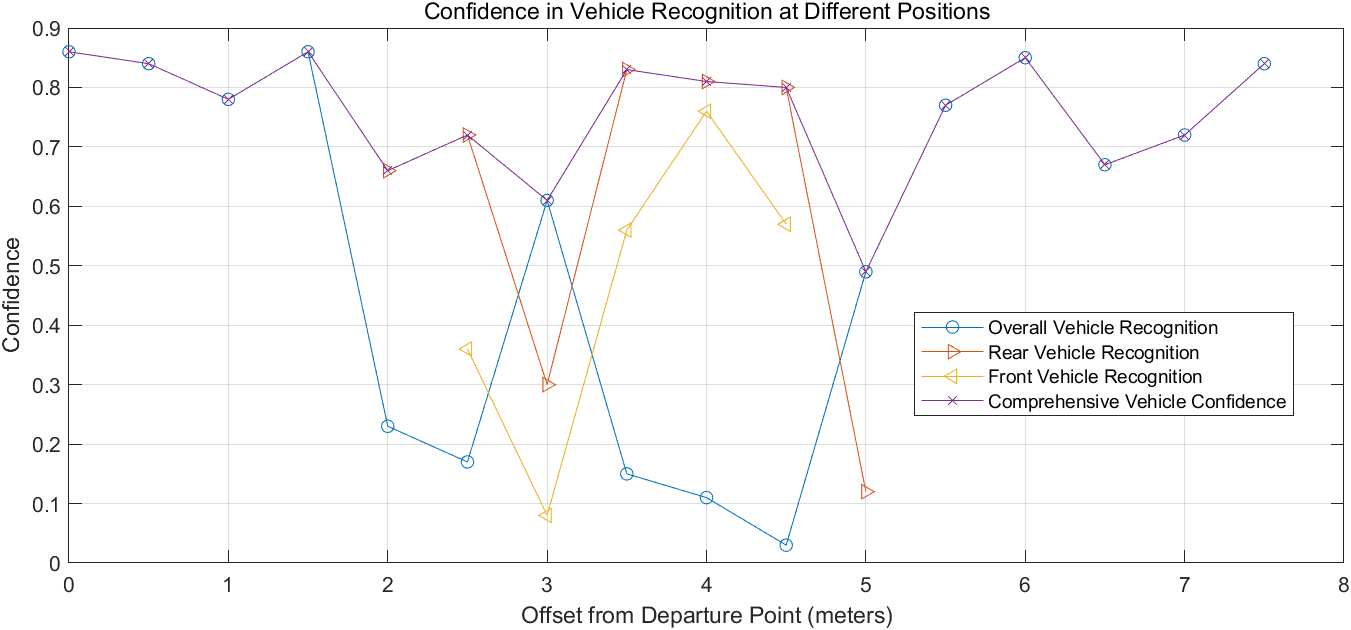}
    \caption{Case1 Recognition Result}
    \label{fig:occ_1_res}
\end{figure*}

In this case, we employ a simulated driving approach to log the recognition confidence of the identified vehicles at various positions. We record the vehicle's position every 50 centimetres and conduct recognition tests. 
The starting and ending points of the vehicle are illustrated in \autoref{fig:occ_1_st} and \autoref{fig:occ_1_ed}. After deriving the recognition confidence of the test subject vehicle at a series of different locations, \autoref{fig:occ_1_res} can be obtained.

\subsubsection{case two}
\begin{figure*} [!ht]  \centering
    \begin{minipage}{0.45\linewidth}       \centering
      \includegraphics[width=\linewidth]{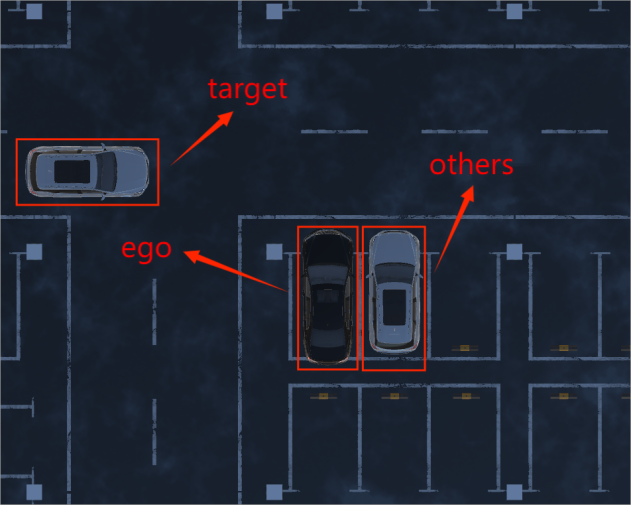}
      \caption{Case2 Start State}
      \label{fig:occ_2_st}
    \end{minipage}
    \begin{minipage}{0.45\linewidth}       \centering
      \includegraphics[width=\linewidth]{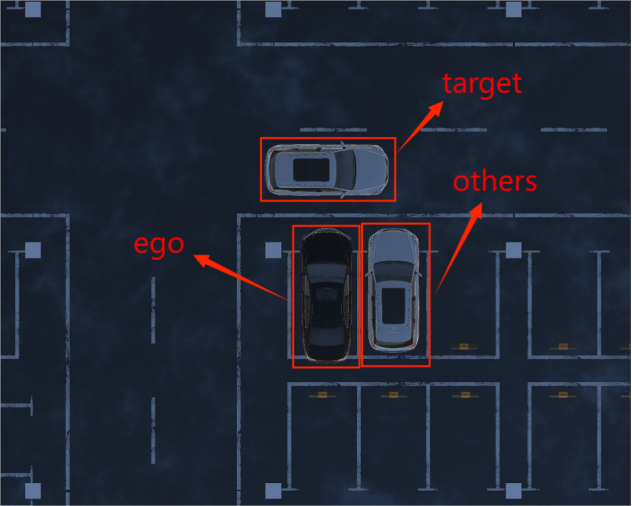}
      \caption{Case2 End State}
      \label{fig:occ_2_ed}
    \end{minipage}
    \hspace{0.05\linewidth}
\end{figure*}

\begin{figure}[!ht]  \centering
    \includegraphics[width=0.47\textwidth]{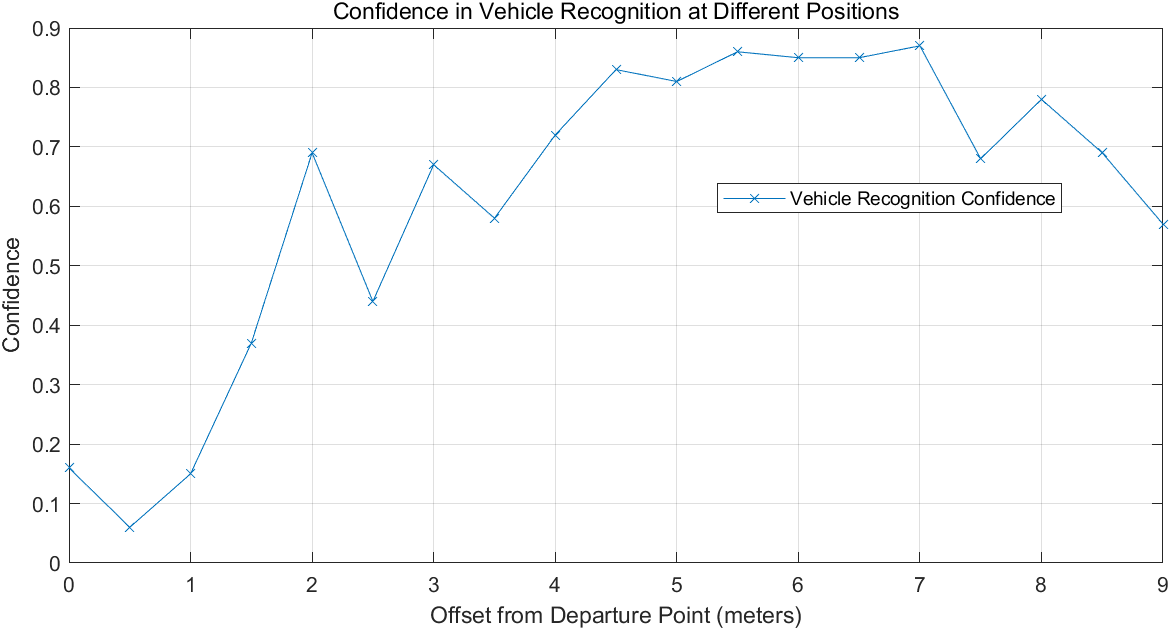}
    \caption{Case2 Recognition Result}
    \label{fig:occ_2_res}
\end{figure}

In contrast to the first case, in this test, the columns are positioned closer to the self-vehicle. The influencing factor here is the layout of the parking spaces, while in the first case, the primary influencing factor was the size of the lane departure warning zone.
We again employ a simulated driving approach to record the recognition confidence of the identified vehicles at various positions. We record the vehicle's position every 50 centimetres and conduct recognition tests. The starting and ending points of the vehicle are illustrated in \autoref{fig:occ_2_st} and \autoref{fig:occ_2_ed}. After deriving the recognition confidence of the test subject vehicle at a series of different locations, \autoref{fig:occ_2_res} can be obtained.

\subsubsection{case three}
\begin{figure*} [!ht]  \centering
    \begin{minipage}{0.45\linewidth}       \centering
      \includegraphics[width=\linewidth]{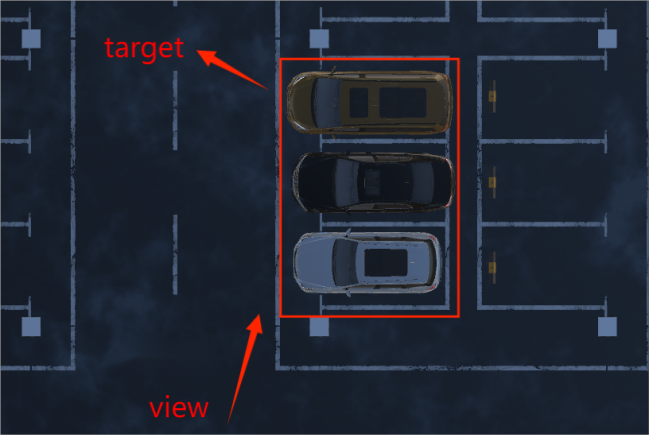}
      \caption{Case3 2D View}
      \label{fig:occ_3_2d}
    \end{minipage}
    \begin{minipage}{0.45\linewidth}       \centering
      \includegraphics[width=\linewidth]{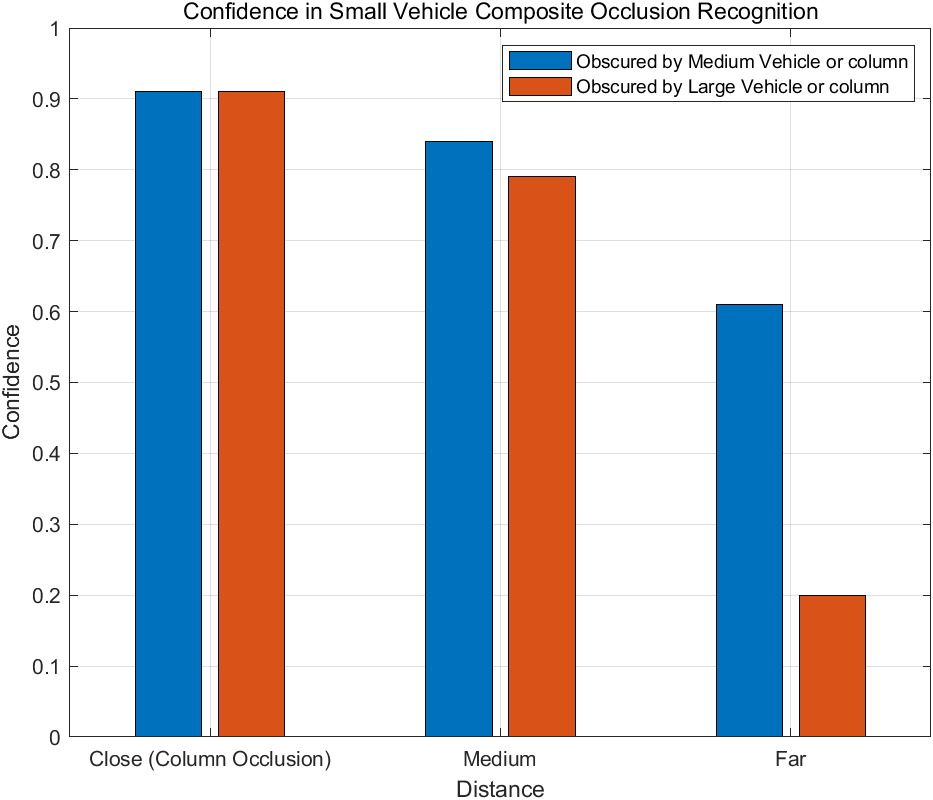}
      \caption{Case3 Result1}
      \label{fig:occ_3_res1}
    \end{minipage}
    \begin{minipage}{0.45\linewidth}       \centering
      \includegraphics[width=\linewidth]{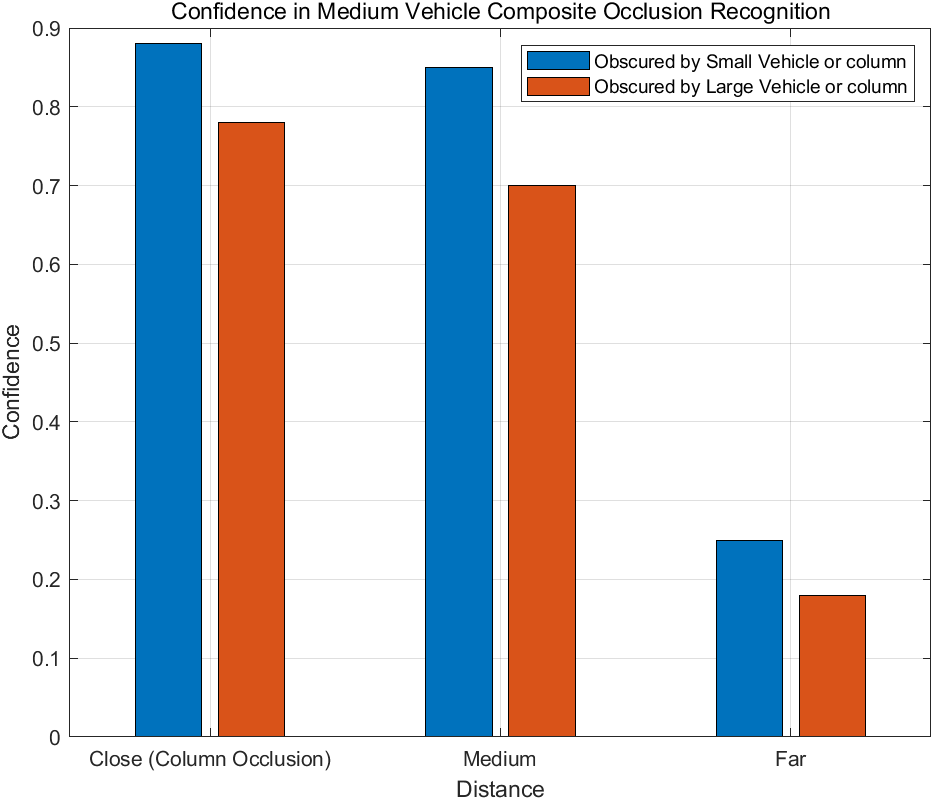}
      \caption{Case3 Result2}
      \label{fig:occ_3_res2}
    \end{minipage}
    \begin{minipage}{0.45\linewidth}       \centering
      \includegraphics[width=\linewidth]{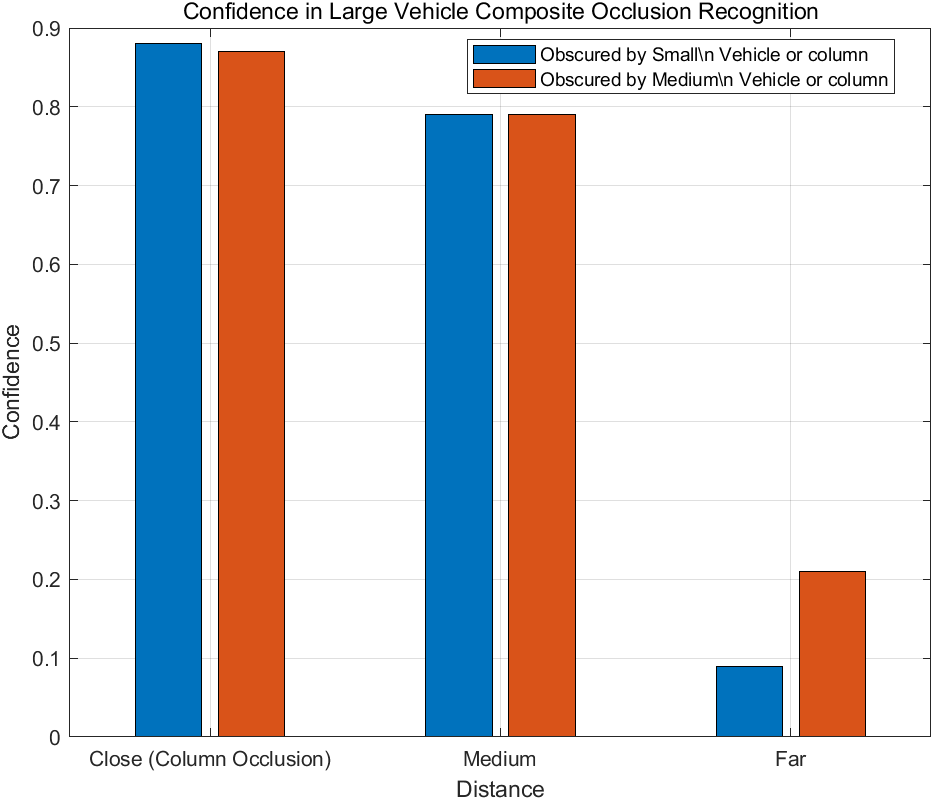}
      \caption{Case3 Result3}
      \label{fig:occ_3_res3}
    \end{minipage}
    \hspace{0.05\linewidth}
\end{figure*}

\begin{figure*}   \centering
    \includegraphics[width=0.95\textwidth]{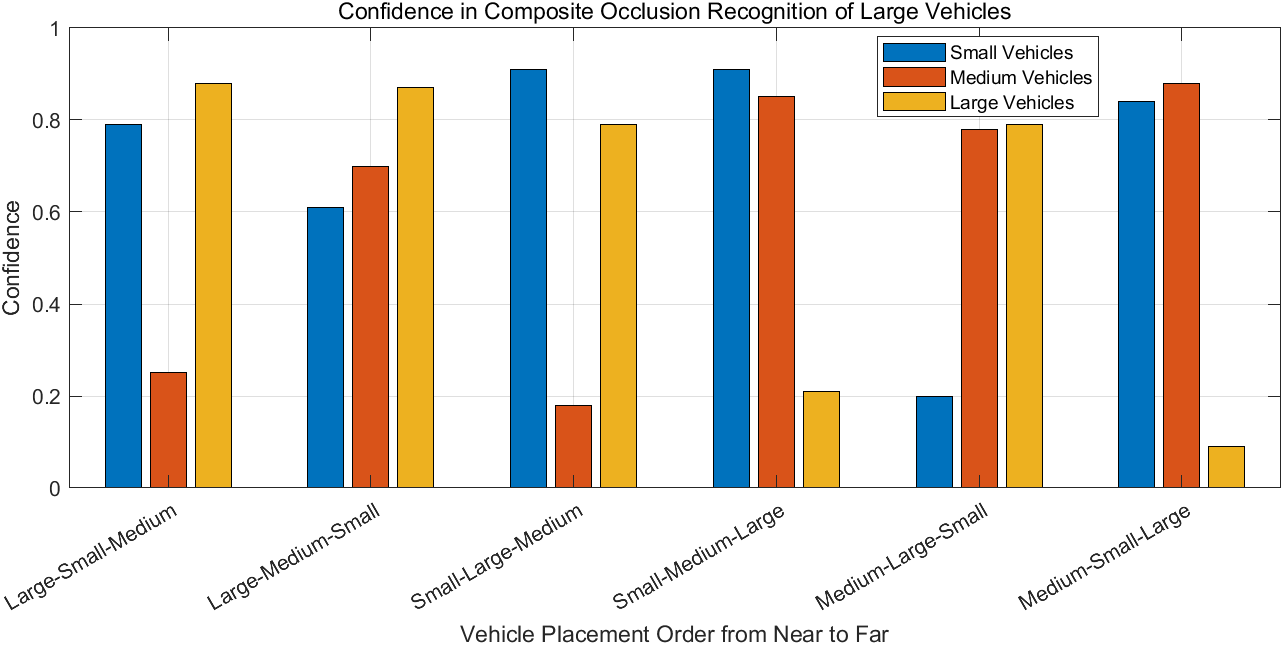}
    \caption{Case3 Result4}
    \label{fig:occ_3_res4}
\end{figure*}

In this case, we selected three different types of vehicles, and different parking positions, as shown in \autoref{fig:occ_3_2d}, according to the occlusion ability from far to near for large, medium, and small vehicles respectively. 
First, we conducted individual tests for three distinct vehicle types placed in different parking positions to assess the recognition confidence of the recognition algorithm. 
The results of these tests are presented in \autoref{tab:occ_3_blank}.
Subsequently, we arranged and combined the vehicles in various configurations within the parking spaces to evaluate the impact of compounded occlusion on recognition confidence. 
The data obtained from these tests are represented in \autoref{fig:occ_3_res1} to \autoref{fig:occ_3_res4}.
\begin{table}[!ht]
    \centering
    \begin{tabular}{c|cccc} \toprule
         & Far & Medium & Close\\ \midrule
        Small vehicle & 0.90 & 0.90 & 0.88\\
        Medium Vehicle & 0.87 & 0.89 & 0.86\\
        Large vehicle & 0.72 & 0.89 & 0.87\\ \bottomrule
    \end{tabular}
    \caption{Control group of Occlusion Case3}
    \label{tab:occ_3_blank}
\end{table}

\section{Discussion}
\label{sec:discussion}
In this experiment, we solely considered the camera as an automated driving vehicle sensor, restricting our analysis to a single position and field of view. 
In reality, most autonomous vehicles equipped with Advanced Driver Assistance Systems (ADAS) feature a variety of sensor types, including multiple cameras with different positions and diverse fields of view.
For instance, forward-facing cameras can be located at the vehicle's top, front, or rearview mirror, each with a field of view ranging from 30 degrees to 120 degrees. 
To better simulate the perception system of vehicle cameras in real-world scenarios, it is advisable to incorporate multiple cameras positioned at different locations with varying fields of view into the experimental setup.
Furthermore, in terms of experimental design, in addition to the three situations mentioned in the experiment, one can also consider introducing more scenarios that challenge the perception algorithm, using recognition objects with varying physical characteristics for multiple iterations.

\section{Conclusions}
\label{sec:conclusion}
From the above experiments, we can draw several conclusions:
\begin{itemize}
    \item Light intensity has virtually no impact on the performance of the YOLO algorithm.
    \item When occlusion is not overly severe, the location of occluded areas affects recognition, but the overall confidence level remains relatively high.
    \item The specific occluded area significantly affects vehicle recognition, with the extent of occlusion having no decisive impact.
    \item In cases of compounded occlusion, recognizing vehicles in more distant parking spaces becomes more challenging, with lower confidence levels when the obstructing vehicle is larger than the target vehicle, making recognition even more difficult.
\end{itemize}

\par
These conclusions imply that "challenging" underground parking scenarios exhibit the following characteristics: dim lighting, substantial occlusion of the field of view by pillars in certain instances, occluding objects covering critical vehicle features in some cases, and the presence of larger vehicles parked in spaces obscuring smaller ones. 
To effectively challenge the lateral AVP algorithm under certain constraints, it is imperative to create underground parking scenarios that encompass various levels of severe occlusion. 
Additionally, the factor of occlusion should be incorporated into the design of the underground parking environment, including the specific arrangement and types of occluding objects within the scene.
We will apply other artificial intelligence technologies such as knowledge graph \cite{liu2022towards,lan2022semantic} and optimization algorithm \cite{lan2022time} to generate scenarios in future work.










\bibliographystyle{IEEEtran}
\bibliography{bibliography}

\end{document}